# A Spatial and Temporal Non-Local Filter Based Data Fusion Method

Qing Cheng[a], Huiqing Liu[b], Huanfeng Shen[b], Penghai Wu[b], Liangpei Zhang[c]

[a] School of Urban Design, Wuhan University, Wuhan, Hubei, 430072, China
[b] School of Resource and Environmental Science, Wuhan University, Wuhan, Hubei, 430079, China
[c] The State Key Laboratory of Information Engineering in Surveying, Mapping and Remote Sensing, Wuhan University, Wuhan, Hubei, 430079, China

Abstract—The trade-off in remote sensing instruments that balances the spatial resolution and temporal frequency limits our capacity to monitor spatial and temporal dynamics effectively. The spatiotemporal data fusion technique is considered as a cost-effective way to obtain remote sensing data with both high spatial resolution and high temporal frequency, by blending observations from multiple sensors with different advantages or characteristics. In this paper, we develop the spatial and temporal non-local filter based fusion model (STNLFFM) to enhance the prediction capacity and accuracy, especially for complex changed landscapes. The STNLFFM method provides a new transformation relationship between the fine-resolution reflectance images acquired from the same sensor at different dates with the help of coarse-resolution reflectance data, and makes full use of the high degree of spatiotemporal redundancy in the remote sensing image sequence to produce the final prediction. The proposed method was tested over both the Coleambally Irrigation Area study site and the Lower Gwydir Catchment study site. The results show that the proposed method can provide a more accurate and robust prediction, especially for heterogeneous landscapes and temporally dynamic areas.

## I. INTRODUCTION

Capturing the spatial and temporal dynamics is a significant issue for many remote sensing based monitoring systems (e.g., the monitoring of land-cover change, intraseasonal ecosystem variations, and atmospheric environment dynamics). However, due to technical limitations, remote sensor designs have a trade-off between the spatial resolution and the revisit cycle [1]–[2], which limits our capacity to acquire remote sensing data with both high spatial resolution and high temporal resolution. For example, the data acquired from the Landsat Thematic Mapper (TM) or Enhanced Thematic Mapper Plus (ETM+) sensors and the SPOT High Resolution Visible (HRV) sensor with a 10–30 m spatial resolution are commonly applied for land-use mapping and biophysical parameter estimation [3]–[6]. However, such data cannot be used to capture rapid surface changes such as crop growth and natural disasters due to their long revisit cycles (Landsat TM/ETM+: 16-day; SPOT HRV: 26-day) and frequent cloud contamination. In contrast, the Terra/Aqua Moderate Resolution Imaging Spectroradiometer (MODIS) and National Oceanic and Atmospheric Administration (NOAA) Advanced Very High Resolution Radiometer (AVHRR) sensors can provide high temporal resolution (daily) observations, and are often applied for monitoring at global scales [7]–[8]. However, data from these sensors cannot be used for research at heterogeneous local scales because of their coarse spatial resolution (250m~1000m). Therefore, combining the advantages of the different sensors by spatiotemporal data fusion methods is considered as a cost-effective way to solve the "spatial-temporal contradiction" problem [9]–[11], thereby enhancing the capability of remote sensing for monitoring land-surface dynamics, especially in rapidly changing areas.

Gao *et al*. [12] developed a spatiotemporal filter based fusion framework, which is called the spatial and temporal adaptive reflectance fusion model (STARFM), for the sake of blending Landsat and MODIS data to produce daily surface reflectance at Landsat spatial resolution and MODIS temporal frequency. The STARFM algorithm generates the fusion data by a weighted sum of the spectrally similar neighboring information from the high spatial resolution images and the high temporal frequency images. The STARFM algorithm has been shown to be a relatively reliable spatiotemporal data fusion approach, and has been widely applied to the investigation of vegetation dynamics [13], the generation of gross primary productivity [14], the analysis of dryland forest phenology [15], the estimation of daily evapotranspiration [16], the examination of virus dissemination [17], and the monitoring of urban heat islands [18].

Furthermore, some improved spatiotemporal filter based algorithms have since been developed. Hilker *et al*. [19] proposed the spatial temporal adaptive algorithm for mapping reflectance change (STAARCH) to identify highly detailed spatial and temporal patterns in land-cover changes. Zhu *et al*. [20] developed an enhanced STARFM model (ESTARFM) to enhance the prediction of the reflectance of heterogeneous landscapes, by assigning different conversion coefficients for homogeneous and heterogeneous pixels. Fu *et al*. [21] modified the procedure of similar pixel selection for the ESTARFM model with an optimal window size and additional ancillary data. Shen *et al*. [22] improved the step of weight calculation for the original STARFM model, by considering sensor observation differences for varied land-cover types. However, this approach requires a prior unsupervised classification for the fine spatial resolution data. Wu *et al*. [23] developed a spatio-temporal integrated temperature fusion model (STITFM) to expand the traditional two-sensor fusion into fusion of data from an arbitrary number of sensors with a unified model.

In other frameworks, Hansen *et al*. [24] used regression trees to integrate Landsat and MODIS data on a 16-day repeat time to monitor forest cover change in the Congo Basin. This method demands a single "best" image to map forest cover status for a given year or decade. Zurta-Milla *et al*. [25] developed an unmixing-based fusion framework to produce Landsat-like images having the spectral and temporal resolution provided by the Medium Resolution Imaging Spectrometer (MERIS). However, this unmixing-based fusion approach requires a prior unsupervised classification for the input fine spatial resolution images, or a high spatial resolution land-use database as auxiliary material for the pixel unmixing. Learning-based spatiotemporal fusion frameworks have been developed in recent years [26]–[27], which are generally based on sparse representation and compressive sensing. This approach can predict both the phenology change and the land-cover type change during an observation period, in a unified way. Nonetheless, the practicability of the learning-based fusion methods needs to be further verified. Moreover, image super-resolution (SR) [28]–[30] can be considered as a different kind of technique to improve the spatial resolution for low spatial resolution but high temporal frequency images.

Generally speaking, the spatiotemporal filter based fusion framework has been the most popular category of spatiotemporal fusion approach so far. Although it has been improved in many different ways, the spatiotemporal filter based fusion framework still has some shortcomings that need to be improved, including the complex change prediction ability and the robustness of the prediction model. In this paper, the spatial and temporal non-local filter based fusion model (STNLFFM) is presented. The main contributions of the proposed STNLFFM method are as follows. 1) Unlike the conventional spatiotemporal filter based fusion algorithms, which focus on the transformation relationship between the fine-resolution data and the coarse-resolution data, the STNLFFM algorithm pays more attention to the relationship between two fine-resolution images acquired from the same sensor at different dates. In the prediction model, it introduces two regression coefficients determined by the reflectance changes from the reference date to the prediction date, to more accurately describe the transformation relationship between the two fine-resolution images, thereby enhancing the prediction capability for complex changed landscapes. 2) Also in the prediction model, STNLFFM introduces the idea of the spatiotemporal non-local filter method, which takes advantage of the highly redundant information of the image sequence. Both the local neighborhood similarity information and the spatiotemporal non-local similarity information are used to jointly produce the unknown pixels at the prediction date, to ensure a more accurate and robust prediction. 3) The proposed STNLFFM method uses a simple and effective weight calculation approach under the non-local filter framework. 4) The STNLFFM method provides a solution for the problem of temporal difference measurement [22] in the procedure of similar pixel selection.

## II. METHOD

### A. The Theoretical Basis of the STNLFFM Method

For convenience, we refer to the image with low spatial resolution but high temporal frequency as the "coarse-resolution" image, while the image with high spatial resolution but low temporal frequency is called the "fine-resolution" image. For a homogeneous coarse-resolution pixel, neglecting the differences in atmospheric correction, we suppose that the changes of reflectance from date $t_k$ to $t_0$ are linear. This assumption is reasonable over a short time period [20]. Thus, the reflectance of the homogeneous coarse-resolution pixel at $t_0$ can be described by the reflectance at $t_k$ as:

$$C(x, y, B, t_0) = a(x, y, B, \Delta t) * C(x, y, B, t_k) + b(x, y, B, \Delta t) \qquad (1)$$

where $C$ denotes the coarse-resolution reflectance, $(x, y)$ is a given pixel location for both coarse-resolution images at two different dates, $B$ is a given band, $\Delta t = t_0 - t_k$, and $a$ and $b$ are coefficients of the linear regression model between the coarse-resolution reflectance at date $t_0$ and $t_k$.

We assume that the coarse-resolution sensor has similar spectral bands to the fine-resolution sensor. We also suppose that the coarse-resolution image has been up-sampled to the same spatial resolution, size, and bounds as the fine-resolution image. Neglecting the geolocation errors and differences in atmospheric correction, the linear relationship between the coarse-resolution reflectance images, as in (1), can also apply to the fine-resolution reflectance images at date $t_0$ and $t_k$. The fine-resolution reflectance at date $t_0$ is then calculated as:

$$F(x,y,B,t_0) = a(x,y,B,\Delta t) * F(x,y,B,t_k) + b(x,y,B,\Delta t) \qquad (2)$$

where $F$ denotes the fine-resolution reflectance. Here, the regression coefficients $a$ and $b$ are determined by the reflectance changes from date $t_k$ to $t_0$. It is notable that coefficients $a$ and $b$ may vary with location due to the complexity of the land cover, and thus they are derived locally rather than using global coefficients.

In fact, the land cover may undergo significant and complex changes during the prediction period. In order to make the prediction more accurate, we take advantage of the same-class pixels (similar pixels) within the image, by considering the same-class pixels with similar reflectance changes over time. As we all know, remote sensing images are often used for generating a wide range of surface information, so there will always be lots of similar information within an image. This similar information not only includes the local neighborhood similarity, but also includes the non-local similarity [31]–[35], such as some repeated ground information, or long edge structures. Moreover, the temporal correlation between remote sensing image sequences makes the similar information (redundancy) even greater, as shown in Fig.

1. These observations prompted us to introduce the idea of non-local filtering [33]–[34], which attempts to make full use of the high degree of redundant information in the image restoration.

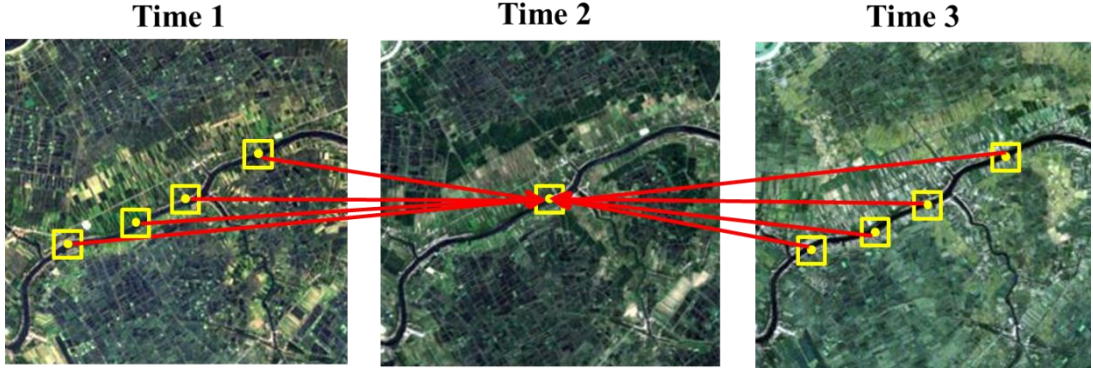

Fig. 1. The spatial and temporal non-local similarity in remote sensing imagery.

B. *Non-Local Filter*

The non-local filter is an effective image denoising algorithm [33]–[35]. Its basic idea is to estimate an unknown pixel with image redundancy. More precisely, given a noisy image $f$, $\Omega$ is its pixel domain. The restored value of a pixel $(x, y)$ in the image $f$ is:

$$NLM_f(x, y) = \sum_{(x_i, y_j) \in \Omega} w_f(x_i, y_j) f(x_i, y_j) \quad (3)$$

where

$$w_f(x_i, y_j) = \frac{1}{C(x, y)} \exp\left\{-\frac{G_a * \|f(P_{(x_i, y_j)}) - f(P_{(x, y)})\|^2}{h^2}\right\} \quad (4)$$

where $G_a$ is the Gaussian kernel with standard deviation $a$; $C(x, y)$ is the normalizing factor; $h$ is the filtering parameter, which is positively related to the noise intensity; and $P_{(x_i, y_j)}$ is a patch centered at point $(x_i, y_j)$. Following (3), the current pixel is restored by averaging the other similar pixels in the image.

*C. The Prediction Model of STNLFFM*

As noted above, the similar information in the images can be used to enhance the reflectance prediction, so we employ the idea of the non-local filter. Moreover, in order to take advantage of the redundant information that exists in both the spatial and temporal directions, we propose the spatial and temporal non-local filter based fusion model (STNLFFM) (5), which integrates the image's spatial and temporal redundancy into the fine-resolution reflectance calculation:

$$F(x, y, B, t_p) = \sum_{k}\sum_{i=1}^{N} W(x_i, y_i, B, t_k) \times [a(x_i, y_i, B, \Delta t_k) * F(x_i, y_i, B, t_k) + b(x_i, y_i, B, \Delta t_k)]$$

(5)

where $F(x, y, B, t_p)$ is the fine-resolution reflectance of the target (predicted) pixel $(x, y)$ at prediction date $t_p$; and $N$ is the number of similar pixels (with the same land-cover type as the target pixel) within the image, including the target pixel itself. $(x_i, y_i)$ is the location of the *i*th similar pixel, and $W(x_i, y_i, B, t_k)$ is the weight of the *i*th similar pixel of the fine-resolution reflectance image at base date $t_k$. It is to note that, a way to more effectively exploit the image redundancy is by searching for similar pixels [35], so we select similar pixels to estimate the target pixel (5), instead of taking all the image pixels to estimate the target pixel, as in the original non-local filtering model (3).

*1) The Weight Calculation:*

The weight $W$ decides the contribution of each similar pixel to predicting the fine-resolution reflectance of the target pixel. According to the non-local filtering framework, the weight $W$ is determined by the reflectance similarity between the similar pixel and the target pixel. Since the fine-resolution reflectance of the target pixel at prediction date $t_p$ is unknown, we use the coarse-resolution reflectance difference between the similar pixel and the target pixel to measure the reflectance similarity, and we then propose a non-local filter based individual weight $W_{individual}$ as follows:

$$W_{individual}(x_i, y_j, B, t_k) = \exp\left(-\frac{G_a * \|C(P(x_i, y_j, B, t_k)) - C(P(x, y, B, t_p))\|}{h^2}\right) \quad (6)$$

where $h$ is the filtering parameter, which is positively related to the intensity of the noise; $G_a$ is the Gaussian kernel with standard deviation $a$; and $C(P(x_i, y_j, B, t_k))$ is the coarse-resolution reflectance of patch $P$ centered at pixel $(x_i, y_j)$. The size of patch $P$ is related to the spatial resolution difference between the coarse- and fine-resolution data input. If the spatial resolution difference is large, it is better to set a small size for patch $P$, since the ground structures in the up-sampled coarse-resolution data might be not that clear and would not benefit the similarity identification. In contrast, if the spatial resolution difference is small, we can set a relatively large size for patch $P$, as the structure information in the up-sampled coarse-resolution data would benefit the similarity identification. The non-local weight $W_{individual}$ determines the weight of the individual pixel. Furthermore, considering that fine-resolution data closer in date to the prediction date should provide more reliable reflectance information, it is reasonable to set a larger weight for the fine-resolution data input in this case. Thus, we introduce the whole weight

$W_{whole}$, which is calculated according to the change magnitude detected by the coarse-resolution reflectance between reference date $t_k$ and prediction date $t_p$, and this weight $W_{whole}$ is applied within each local window $w \times w$ (7), to decide which fine-resolution image input provides more reliable information in the local window:

$$W_{whole} = \frac{1/\sum_{i=1}^{w}\sum_{j=1}^{w}(|C(x_i,y_j,B,t_k)-C(x_i,y_j,B,t_p)|)}{\sum_{k}(1/\sum_{i=1}^{w}\sum_{j=1}^{w}(|C(x_i,y_j,B,t_k)-C(x_i,y_j,B,t_p)|))} \quad (7)$$

A larger value of $W_{whole}$ means that the fine-resolution reflectance at date $t_k$ should be given a higher weight. Then, synthesizing the two weights $W_{individual}$ and $W_{whole}$, the final weight of the similar pixel $(x_i, y_j)$ in the fine-resolution reflectance image at reference date $t_k$ is:

$$W(x_i, y_j, B, t_k) = W_{individual}(x_i, y_j, B, t_k) * W_{whole} \quad (8)$$

*2) The Regression Coefficients Calculation:*

The regression coefficients $a$ and $b$ for each similar pixel in the fine-resolution reflectance images can be calculated from the available coarse-resolution reflectance images. Since the similar pixels have the same reflectance change as the target pixel, they should have the same regression coefficients. Thus, it is feasible to make use of the information from similar pixels to compute the regression coefficients. In the case of the coarse-resolution reflectance at dates $t_k$ and $t_p$ being perfectly correlated, i.e., the regression coefficient $a$ is equal to 1, we have:

$$C(x, y, B, t_p) = C(x, y, B, t_k) + b(x, y, B, \Delta t) \quad (9)$$

The predicted fine-resolution reflectance is then:

$$F(x,y,B,t_p) = F(x,y,B,t_k) + C(x,y,B,t_p) - C(x,y,B,t_k) \qquad (10)$$

Therefore, we can see that the STARFM model is a special case of the STNLFFM model. In reality, due to the complexity of the land cover, the regression coefficient $a$ may not be equal to 1, but varies in the vicinity of 1. Therefore, we apply a restricted least-squares model to the coarse-resolution reflectance of the similar pixels to obtain the regression coefficients $a$ and $b$ for the target pixel:

$$\text{set } A = \begin{pmatrix} a \\ b \end{pmatrix}, I = (1,0)$$

$$\arg\min f(A) = \frac{1}{2} \left[ \begin{pmatrix} C_{p1} \\ C_{p2} \\ \vdots \\ C_{pn} \end{pmatrix} - \begin{pmatrix} C_{k1} & 1 \\ C_{k2} & 1 \\ \vdots & \vdots \\ C_{kn} & 1 \end{pmatrix} * A \right]^2 + \frac{1}{2}\gamma [I * A - 1]^2 \qquad (11)$$

where $C_{pn}$ and $C_{kn}$ are the coarse-resolution reflectance of the $n$th similar pixel at dates $t_p$ and $t_k$, respectively; and $\gamma$ is a regularization parameter.

*3) Similar Pixel Selection:*

As noted before, the pixels with the same land-cover type as the target pixel are the similar pixels. Selecting the similar pixels ensures that the appropriate reflectance information is used for the production of the target fine-resolution pixel, which can avoid trivial calculation and improve the prediction accuracy. Since the similar pixels have close reflectances and changes, we use two constraints (spectral consistency and change consistency) to pick them out. A good candidate pixel $(x_i, y_j)$ should satisfy the following conditions for the reflectance of all the bands:

$$|F(x_i, y_j, B, t_k) - F(x, y, B, t_k)| \leq d * 2^{F(x,y,B,t_k)} \qquad (12)$$

$$\left\| |C(x_i, y_j, B, t_k) - C(x_i, y_j, B, t_p)| - |C(x, y, B, t_k) - C(x, y, B, t_p)| \right\| < \sigma_{CC} \qquad (13)$$

where $d$ is a free parameter [22]; and $\sigma_{CC}$ is the uncertainty for the temporal difference between the two coarse-resolution reflectance images, which is caused mainly by the bias in the atmospheric correction. Equation (12) is the spectral consistency condition to ensure that the reflectance between a similar pixel and the target pixel is close. It is noteworthy that (13) is the change consistency condition, which means that a pixel whose temporal change between date $t_k$ and prediction date $t_p$ is close to that of the target pixel is more likely to be selected as a similar pixel. For the STARFM method [12], it regards a pixel whose temporal change between dates $t_k$ and $t_p$ is smaller than that of the target pixel as a candidate similar pixel. However, this will result in a large error when the predicted target pixel shows considerable phenological change, because a pixel with the same land-cover type would also show considerable reflectance change in this case. There are also some other papers [13], [22] in which the temporal difference measurement is excluded in the procedure of selecting similar pixels, for the sake of avoiding the error mentioned above. Here, we use (13) to effectively solve the problem of temporal difference measurement and improve the accuracy of selecting similar pixels. In practice, we choose a searching window centered at the target pixel, as in [34], since searching for the similar pixels over the whole image domain is very expensive.

### D. Implementation Process of STNLFFM

In practice, we use at least two pairs of fine- and coarse-resolution images acquired at the reference dates ($t_m$ and $t_n$) and a coarse-resolution image acquired at the prediction date ($t_p$) to predict the desired fine-resolution image. The STNLFFM algorithm implementation includes three main steps: 1) for each pixel in the reference fine-resolution images, we search for pixels similar to it in the image; 2) the weights $W$ of all the similar pixels are calculated; and 3) the regression coefficients $a$ and $b$ for each similar pixel are calculated from the available coarse-resolution reflectance images. Finally, as in (5), the weights and regression coefficients are applied to the two fine-resolution reflectance images acquired at the reference dates to produce the fine-resolution reflectance at the desired prediction date. All of the steps are discussed in detail below. The flowchart of the STNLFFM method is presented in Fig. 2.

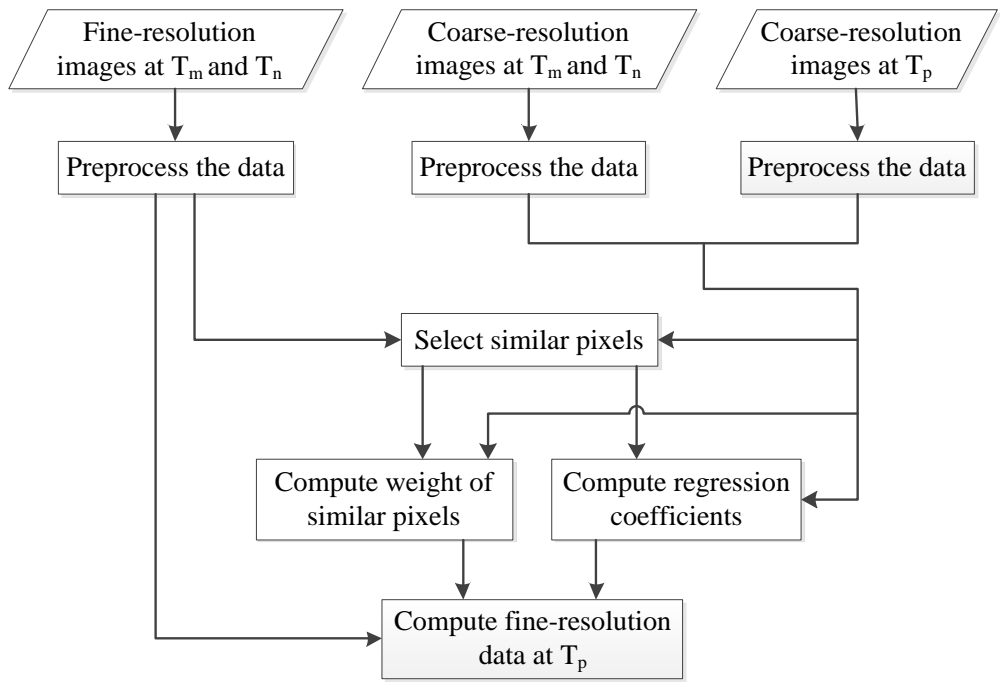

Fig. 2. The flowchart of the STNLFFM algorithm.

*E. Differences Between the STNLFFM Method and the STARFM/ESTARFM Methods*

Although they all belong to the filter-based fusion framework, the proposed STNLFFM method and the conventional STARFM or ESTARFM methods have big differences, including the form of the prediction model, the method of weight calculation, and the procedure of searching for similar pixels.

*1) Differences in the Construction of the Prediction Models:*

For the prediction model, the proposed STNLFFM uses two regression coefficients *a* and *b* to more accurately describe the reflectance change information, thereby enhancing the prediction capacity for complex changed landscapes. For the STARFM and ESTARFM methods, they focus on the transformation relationship between the fine-resolution image and the coarse-resolution image, and they do not pay enough attention to the relationship between the two fine-resolution images acquired from the same sensor at different times (in other words, the reflectance changes over time are not sufficiently considered). Thus, they suppose that the predicted high-resolution image is perfectly correlated with the reference high-resolution image (with the ratio equal to 1) in their prediction models, and, as noted before, the STARFM model is a special case of the STNLFFM model with regression coefficient $a = 1$. That is to say, the STARFM/ESTARFM models assume that the reflectance changes are perfectly linear over time. However, this assumption may not be appropriate for complex changed landscapes. In contrast, the proposed STNLFFM model introduces two regression coefficients *a* and *b* determined by the reflectance changes from the reference date to the prediction date, to more accurately describe the relationship between the predicted high-resolution image and the reference high-resolution image, thereby improving the prediction accuracy, especially for complex changed landscapes.

Moreover, in the prediction model, STNLFFM employs the idea of non-local filtering to take advantage of the redundant information in the remote sensing image sequence. Both the local neighborhood similarity information and the non-local similarity information are used to jointly produce the unknown pixels. As for the STARFM and ESTARFM methods, they only use the neighborhood similar pixels to predict the unknown pixels. However, for the ground objects with small areas, or edge structures, the number of neighborhood similar pixels may be small, and not enough to provide accurate predictions. The proposed STNLFFM takes advantage of the large amount of redundant information in both the spatial and temporal directions of the data to ensure a more accurate and robust prediction.

2) *Differences in the Weight Calculation:*

STNLFFM separates the weight calculation into individual weight and whole weight calculation. The individual weight is used to measure which similar pixel within a fine-resolution image provides more reliable information; the whole weight is used to measure which fine-resolution image input provides more reliable information. For the individual weight calculation, the original STARFM algorithm takes the spectral difference, the temporal difference, and the location distance into consideration. However, the temporal difference can bring large errors to the problem of temporal difference measurement mentioned above; and the location distance and spectral difference are not that meaningful for the proposed STNLFFM algorithm.

Therefore, according to the non-local filtering method, the STNLFFM algorithm uses the reflectance similarity of image patches to calculate the individual weight, which avoids trivial calculation, improves the calculation efficiency, and can reduce the interference of image noise.

Furthermore, in the procedure of selecting similar pixels, the STNLFFM algorithm resolves the problem of temporal difference measurement. In the STARFM method, a pixel whose temporal change between dates $t_k$ and $t_p$ is smaller than that of the target pixel is selected as a candidate similar pixel. However, this will result in large errors when the predicted target pixel has a considerable phenological change. The STNLFFM algorithm selects a pixel whose temporal change between date $t_k$ and prediction date $t_p$ is closer to that of the target pixel as a candidate similar pixel, which is more reasonable, and can significantly improve the accuracy of selecting similar pixels.

## III. EXPERIMENTS

*A. Study Sites and Data*

The study sites and data tested in this paper are the same as those used in the research of Emelyanova *et al*. [36]. The first study site is the Coleambally Irrigation Area ("CIA" herein) located in southern New South Wales, for which 17 cloud-free Landsat-MODIS pairs are available for the 2001–2002 austral summer growing season. The other study site is the Lower Gwydir Catchment ("LGC" herein) located in northern New South Wales, for which 14 cloud-free Landsat-MODIS pairs are available from April 2004 to April 2005. All the Landsat images were atmospherically corrected in the same way as the research of Emelyanova *et al*. [36]: the CIA images were atmospherically corrected using MODTRAN4 [37], and the LGC images were atmospherically corrected using Li *et al*.'s algorithm [38]. For both study sites, the latest MODIS Terra MOD09GA Collection 6 data were used. These data were up-sampled to the same spatial resolution (25 m) as the Landsat data using a cubic convolution algorithm.

The CIA has an overall area of 2193 km$^2$ (1720 × 2040 pixels in Landsat images). The irrigation area, which is scattered over the whole study site, exhibits temporal dynamics associated with crop phenology over a single growing season. However, the surrounding agricultural and woodland areas vary less over time. Due to the small field sizes and sporadic distribution of the irrigation area, the CIA can be considered a spatially heterogeneous site. The LGC site has an overall area of 5440 km$^2$ (3200 × 2720 pixels in Landsat images). The temporal extent of the LGC data is approximately 1 year. A large flood occurred in mid-December 2004, which caused inundation over large areas (about 44%). Due to the flooding event leading to different spatial and temporal variations, the LGC is considered a temporally dynamic site. Some of the temporal data from the two sites are shown in Fig. 5.

*B. Quantitative Evaluation Indices*

A Landsat-like image on a certain date is predicted using two Landsat-MODIS pairs on other dates and a MODIS image on the prediction date, as shown in detail below. The predicted Landsat image is compared with the real Landsat image acquired at the prediction date. (The real Landsat images at the prediction date are not used as input, and are only used for the validation.) The RMSE (root-mean-square error) and $R^2$ (determination coefficients) are used to give a quantitative evaluation of the experimental results. The definitions of these evaluation indices are as follows:

$$RMSE = \sqrt{\frac{\sum_{i=1}^{N}(P_i - O_i)^2}{N}} \tag{14}$$

$$R^2 = \left( \frac{\sum_{i=1}^{N}(P_i - \bar{P})(O_i - \bar{O})}{\sqrt{\sum_{i=1}^{N}(P_i - \bar{P})^2} \sqrt{\sum_{i=1}^{N}(O_i - \bar{O})^2}} \right)^2 \tag{15}$$

where $N$ is the total number of pixels in the predicted image; $P_i$ and $O_i$ are the value of the $i$th pixel in the predicted image and the observed image, respectively; and $\bar{P}$ and $\bar{O}$ are the mean values. A smaller RMSE and larger $R^2$ mean that the predicted result is closer to the observed images (i.e., the predicted result is more accurate).

## C. Experimental Results

To verify the efficacy of the STNLFFM algorithm, we conducted two groups of experiments: 1) short time series fusion experiments; and 2) long time series fusion experiments.

*1) Short Time Series Fusion Experiments:*

For both study sites, all the Landsat-MODIS pairs were arranged in chronological order. The STNLFFM algorithm was tested by predicting a Landsat-like image on a certain date using two Landsat-MODIS pairs that were the nearest temporal neighbors to the predicted date, one before and one after, and the MODIS image on the predicted date was also used as an input. All possible combinations of predictions were processed at both study sites. For the CIA site, Landsat-like images at 15 central dates were predicted, and for the LGC site, Landsat-like images at 12 central dates were predicted, except for the first and last dates.

To make a comparative analysis, the STNLFFM method was compared with two popular methods: STARFM and ESTARFM. The mean RMSE and $R^2$ values of all bands for the results of all the prediction dates for the CIA site and LGC site are shown in Fig. 3 and Fig. 4, respectively. We can see from Fig. 3 and Fig. 4 that, in the vast majority of the 15 predicted results of the CIA site and the 12 predicted results of the LGC site, the RMSE values obtained using the STNLFFM method are the lowest, and the $R^2$ values using the STNLFFM method are the highest. That is to say, the proposed STNLFFM method is able to provide a more accurate and robust prediction result than the other two methods.

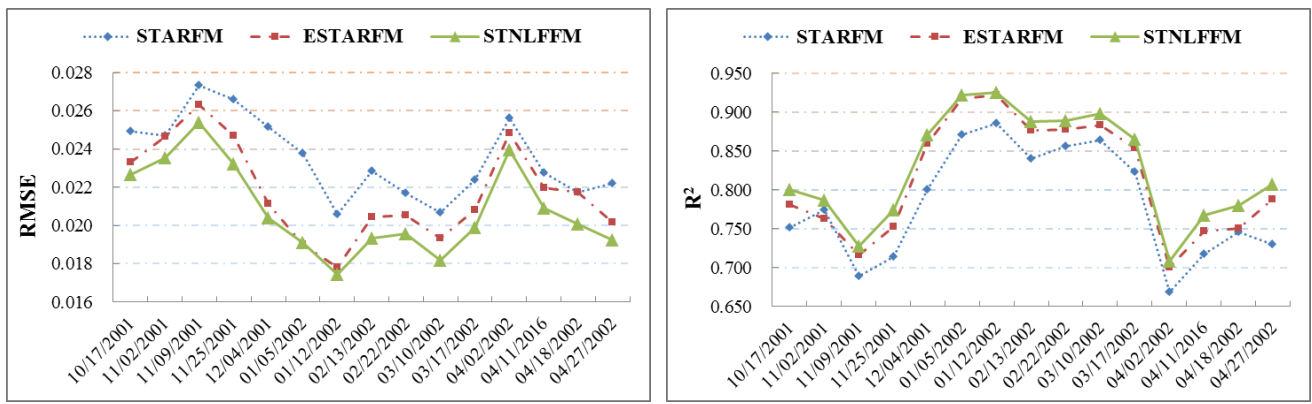

Fig. 3. The mean RMSE and $R^2$ values of all bands for the results of all the prediction dates at the CIA site.

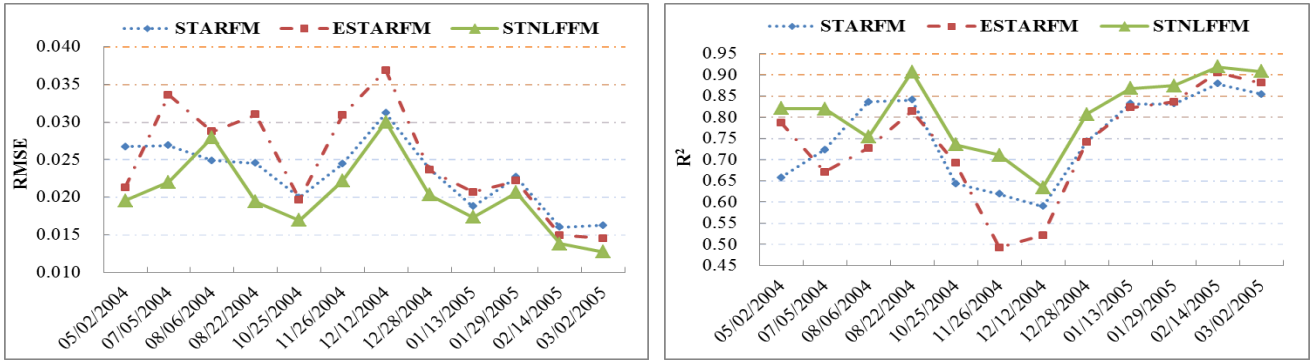

Fig. 4. The mean RMSE and $R^2$ values of all bands for the results of all the predicted dates at the LGC site.

We also present some details of the test data and the above results. Fig. 5 shows the observed Landsat-MODIS pairs on a key date and the two nearest dates at the CIA and LGC sites, respectively. For the CIA site, the images are presented as Landsat bands 4, 3, and 2 (MODIS 2, 1, 4), displayed as RGB, as shown in Fig. 5(a). We can see from Fig. 5(a) that the crop in the sporadic irrigation fields begins to turn green through January to February, but the surrounding agricultural and woodland areas show less change during this time. This makes the CIA area spatially heterogeneous. For the LGC site, the large flood occurred in mid-December 2004, causing temporal dynamics and abnormal change of the ground surface. In order to show the ground features of the LGC site more clearly, the LGC images are presented as Landsat bands 5, 4, and 3 (MODIS 6, 2, 1), displayed as RGB, as shown in Fig. 5(b). We use the Landsat-MODIS pair dates before and after to predict a Landsat-like image at the middle date. The prediction results for these two groups of data are shown in Fig. 6 and Fig. 7.

For the results of the CIA study site (Fig. 6), it can be seen that the three methods are generally all able to predict the crop phenological changes, and the results are satisfactory in most regions. However, for some special heterogeneous regions, such as the zoomed detailed regions in Fig. 6, we can see that, for the STARFM and ESTARFM methods, they produce spectral distortion in their results (Fig. 6(b) and 6(c)), especially for the STARFM method, which is poor at handling spatially heterogeneous areas. For the proposed STNLFFM method, it obtains a visually convincing result which is closest to the observed Landsat data, as shown in Fig. 6(d). For the test results of the LGC study site (Fig. 7), we can see from the zoomed detailed regions that there is some noise arising in the edge regions for the STARFM method (Fig. 7(b)). For the result of the ESTARFM method (Fig. 7(c)), serious spectral distortion occurs in the flooded area since ESTARFM is less good at handling temporally dynamic areas [36]. The proposed STNLFFM method obtains a result (Fig. 7(d)) that is visually similar to the observed Landsat data. To give a quantitative assessment of these results, the RMSE and $R^2$ values of each band in Figs. 6 and 7 are shown in Tables I and II, respectively. It can be seen from Tables I and II that, although the RMSEs obtained in a few bands with the STNLFFM method are not lower than the RMSEs obtained with the ESTARFM method, most of the results obtained using STNLFFM have the lowest RMSE and highest $R^2$. These improvements are clearly evident, especially for the results in Fig. 7 (Table II).

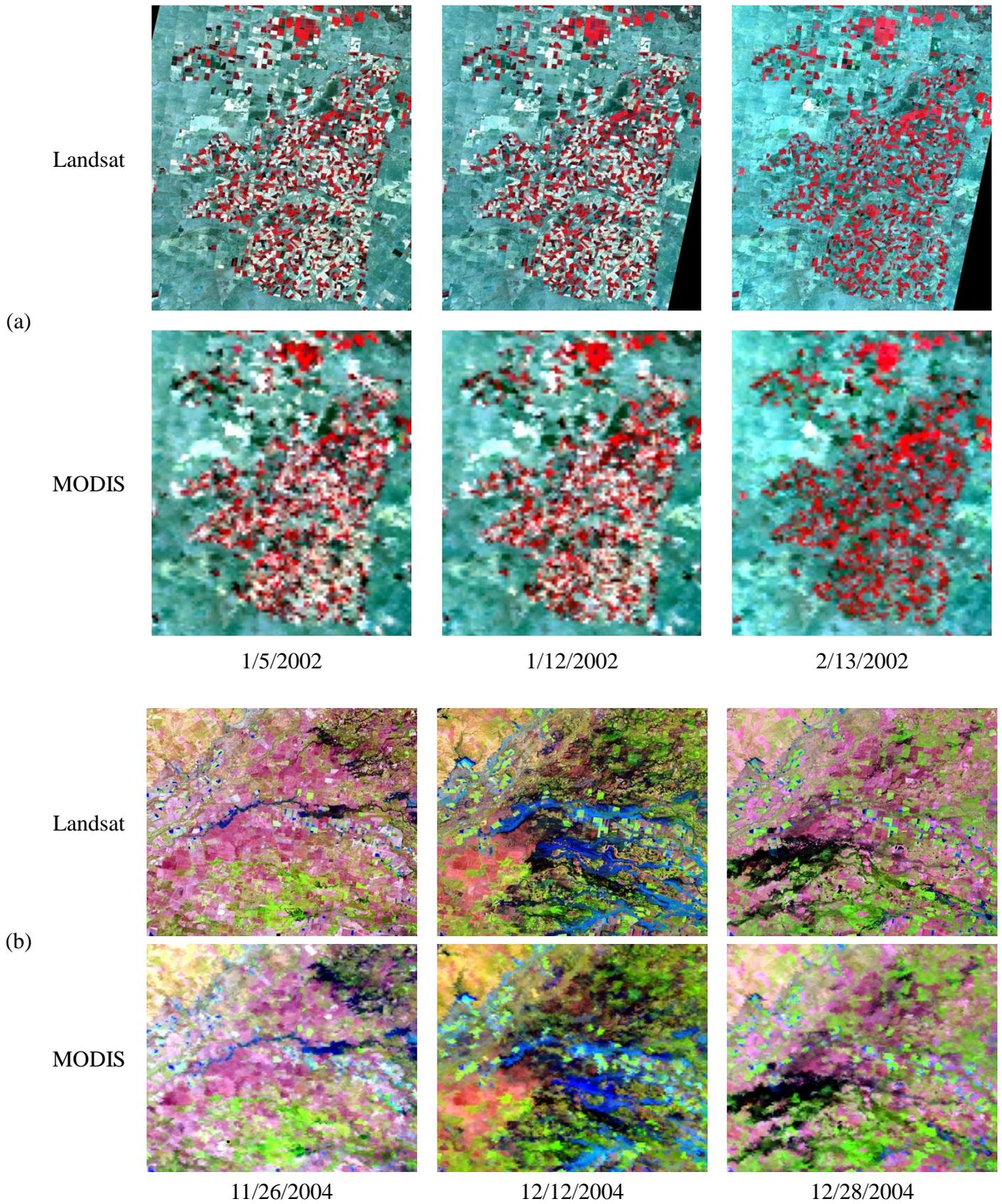

Fig. 5. The observed Landsat-MODIS pairs on a key date and the two nearest dates at the CIA and LGC sites: (a) the observed Landsat-MODIS pairs at the CIA site on January 5, 2002, January 12, 2002, and February 13, 2002, respectively; (b) the observed Landsat-MODIS pairs at the LGC site on November 26, 2004, December 12, 2004, and December 28, 2004, respectively.

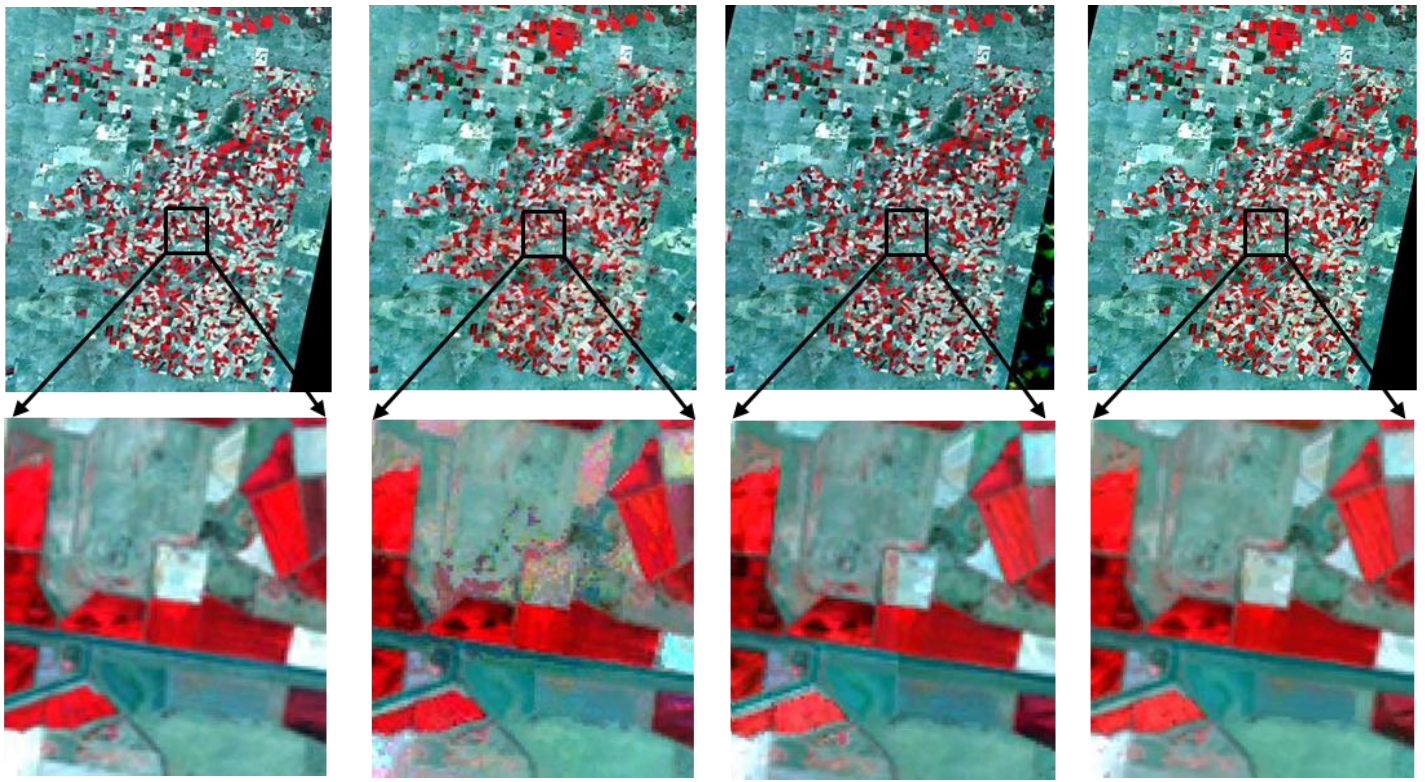

(a) Observation      (b) STARFM      (c) ESTARFM      (d) STNLFFM

Fig. 6. The prediction results of the CIA site on January 12, 2002: (a) the observed Landsat image; (b)–(d) the Landsat-like images predicted by STARFM, ESTARFM, and the proposed STNLFFM method, respectively.

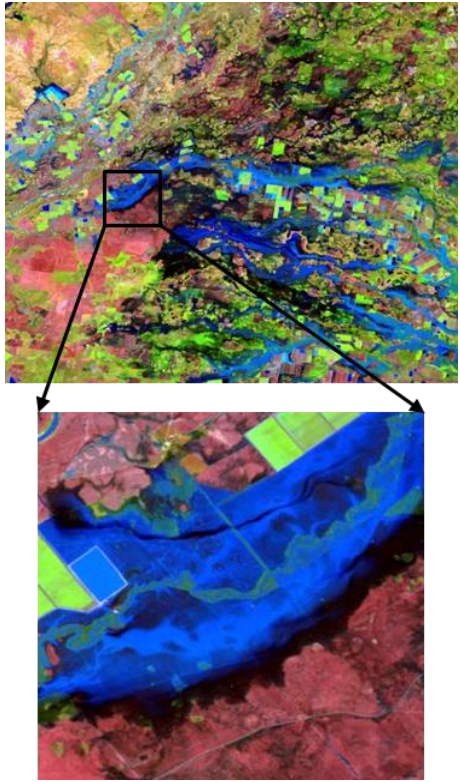
(a) observation

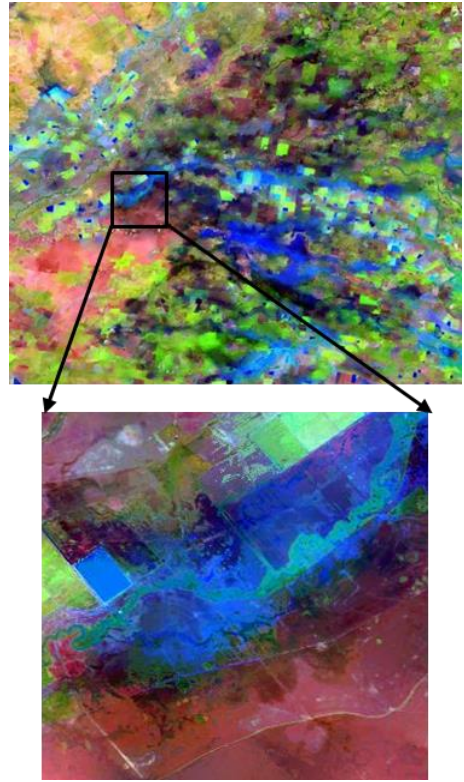
(b) STARFM

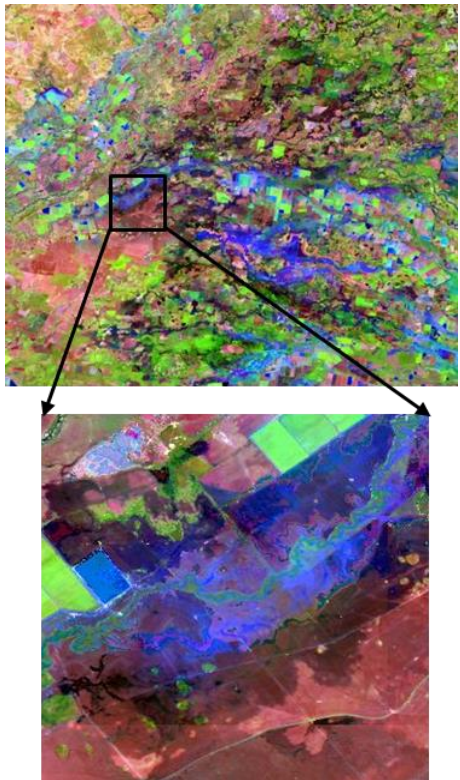
(c) ESTARFM

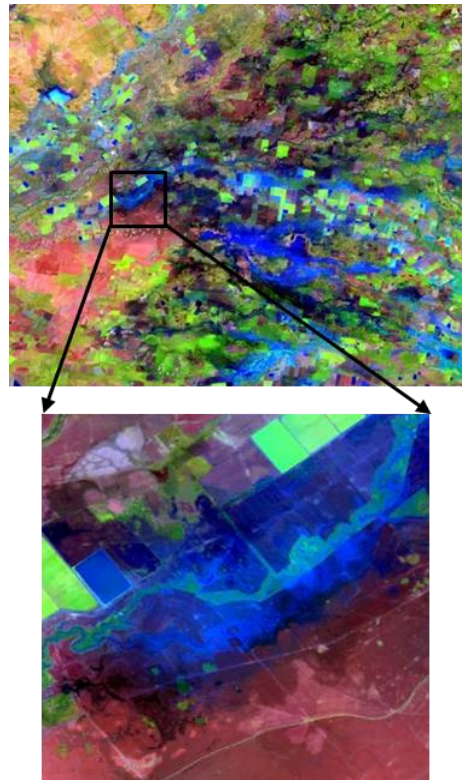
(d) STNLFFM

Fig. 7. The prediction results of the LGC site on December 12, 2004: (a) the observed Landsat image; (b)–(d) the Landsat-like images predicted by STARFM, ESTARFM, and the proposed STNLFFM method, respectively.

TABLE I

The RMSE and $R^2$ values of the prediction results in Fig. 6.

| Method / Band | RMSE | | | R2 | | |
|---|---|---|---|---|---|---|
| | STARFM | ESTARFM | STNLFFM | STARFM | ESTARFM | STNLFFM |
| 1 | 0.0102 | 0.0098 | **0.0093** | 0.88 | 0.91 | **0.93** |
| 2 | 0.0131 | **0.0104** | 0.0104 | 0.88 | 0.92 | **0.93** |
| 3 | 0.0213 | **0.0162** | 0.0166 | 0.89 | **0.94** | 0.94 |
| 4 | 0.0283 | 0.0227 | **0.0218** | 0.81 | 0.88 | **0.89** |
| 5 | 0.0270 | **0.0249** | 0.0249 | 0.92 | **0.93** | 0.93 |
| 6 | 0.0236 | 0.0229 | **0.0216** | 0.92 | **0.93** | 0.93 |

TABLE II

The RMSE and $R^2$ values of the prediction results in Fig. 7.

| Method / Band | RMSE | | | R2 | | |
|---|---|---|---|---|---|---|
| | STARFM | ESTARFM | STNLFFM | STARFM | ESTARFM | STNLFFM |
| 1 | 0.0138 | 0.0148 | **0.0135** | 0.54 | 0.55 | **0.58** |
| 2 | 0.0198 | 0.0203 | **0.0189** | 0.51 | 0.52 | **0.56** |
| 3 | 0.0246 | 0.0255 | **0.0236** | 0.53 | 0.53 | **0.57** |
| 4 | 0.0346 | 0.0408 | **0.0317** | 0.70 | 0.59 | **0.77** |
| 5 | 0.0535 | 0.0631 | **0.0516** | 0.64 | 0.53 | **0.68** |
| 6 | 0.0409 | 0.0567 | **0.0407** | 0.62 | 0.40 | **0.64** |

*2) Long Time Series Fusion Experiments:*

In this part, we analyze the influence of the time interval length on the prediction results. The dates of the LGC image series are irregular and have quite different time intervals, while the dates of the CIA image series are regular and have similar time intervals. In order to make a convenient and effective analysis, we only used the CIA image series in the experiments. The Landsat-MODIS pairs of the CIA site were arranged in chronological order, as before. The middlemost date, 2/13/2002, was treated as the prediction date. The proposed STNLFFM algorithm was tested by predicting a Landsat-like image on the prediction date (2/13/2002) using two Landsat-MODIS pairs on reference dates that were symmetrically distributed with the prediction date, one before and one after, as shown in Fig. 8. As the time intervals between the prediction date and the reference dates were increased, all possible combination of predictions were processed. The RMSE values of the predicted results were calculated, as shown in Fig. 9. The horizontal axis in Fig. 9 represents the average time interval between the prediction date and the two reference dates. From Fig. 9, we can observe the following phenomena:

1) For all three methods, the shorter the time interval, the better the prediction, and vice versa. When the average time interval was longer than about 90 days, the prediction accuracy fluctuated less.

2) From the two curves of STARFM and ESTARFM, we can see that when the average time interval was short (less than about 65 days), the prediction of ESTARFM was better than that of STARFM. This is because the spatial variability is the dominant factor affecting the prediction when the time interval is short, and ESTARFM is better designed to deal with spatial variability than STARFM [20]; however, when the average time interval is long (more than about 65 days), the prediction of STARFM is better than that of ESTARFM. This is because the temporal variability becomes the dominant factor affecting the prediction when the time interval is sufficiently long, and STARFM is better able to deal with temporal variability than ESTARFM [36]. Moreover, we can infer that the intersection of the two curves (STARFM and ESTARFM) is the point where the levels of spatial and temporal variability are equal, and this point is located at approximately 65 days (the average time interval between the prediction date and the two reference dates) for the CIA site.

3) For every prediction result in Fig. 9, the RMSE value of the STNLFFM method is lower than those of the other two methods, which suggests that, with the interference from both the spatial variability and temporal variability, the STNLFFM method can still obtain better prediction results than the other two methods, and the superior performance of the STNLFFM algorithm continues with the increase of the prediction period.

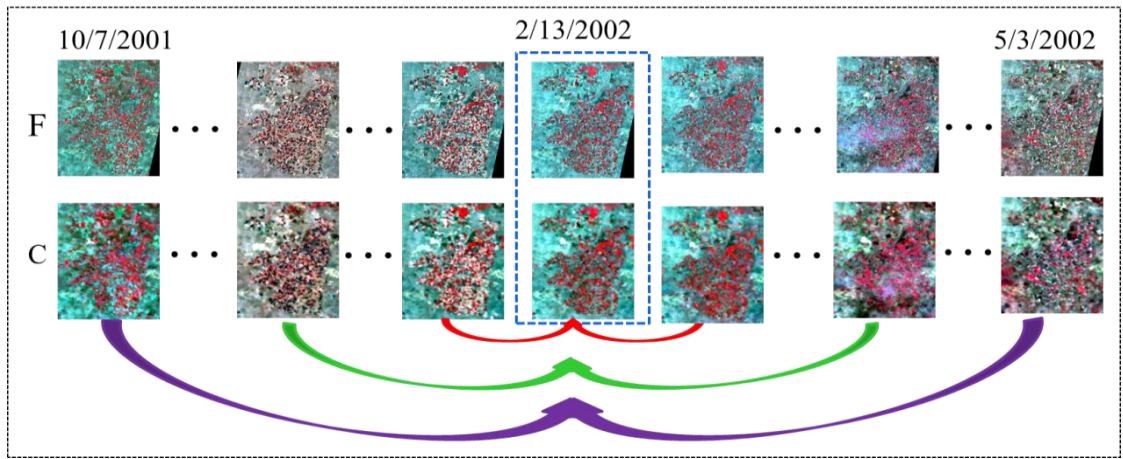

Fig. 8. Predicting a Landsat-like image on the middlemost date (2/13/2002) using two Landsat-MODIS pairs on reference dates that are symmetrically distributed, one before and one after.

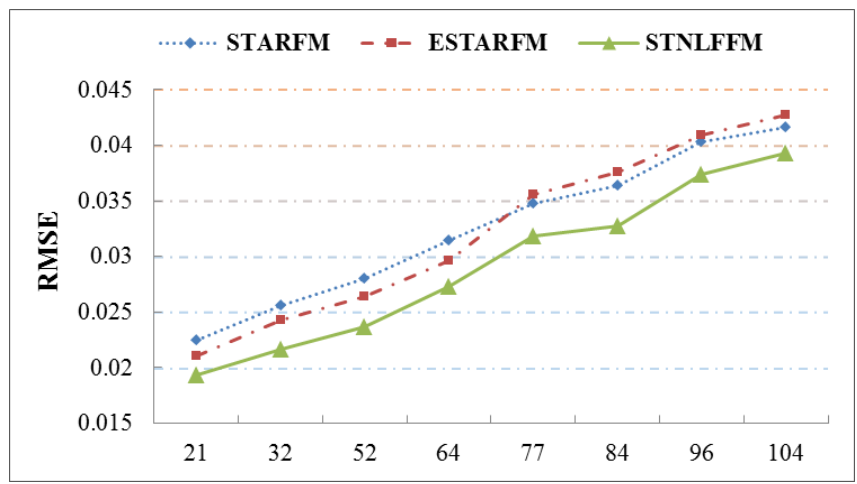

Fig. 9. The RMSE values of the predicted results. The horizontal axis represents the average time interval between the prediction date and the two reference dates.

## IV. CONCLUSION

For the sake of obtaining remote sensing data with both high spatial resolution and high temporal frequency, in this paper, we have proposed the spatial and temporal non-local filter based fusion model (STNLFFM). The STNLFFM algorithm was tested over the Coleambally Irrigation Area study site and the Lower Gwydir Catchment study site, and we conducted two groups of experiments to verify the efficacy of the STNLFFM algorithm. The experimental results show that the STNLFFM algorithm can predict the fine-resolution reflectance accurately and robustly, for both heterogeneous landscapes and temporally dynamic areas. Moreover, the superior performance of the proposed algorithm continues with the increase of the prediction period.

The proposed STNLFFM algorithm makes several improvements to the STARFM and ESTARFM algorithms. Firstly, in the prediction model, the STNLFFM algorithm uses two regression coefficients to more accurately describe the land-cover change information, thereby enhancing the prediction capability for complex changed landscapes. Secondly, STNLFFM introduces the idea of non-local filtering, which takes advantage of the high degree of redundancy in the image sequence to produce a more accurate and robust prediction. Thirdly, STNLFFM uses a simple method of weight calculation which can improve the computational efficiency and reduce the interference of image noise. Lastly, STNLFFM solves the problem of temporal difference measurement in the procedure of searching for similar pixels, and improves the accuracy of similar pixel selection.

There are, however, some limitations to the STNLFFM method. STNLFFM is based on the assumption that the reflectance change rate is linear. However, this assumption might not be appropriate in some situations, especially over a long time period. In addition, although the current computation speed of the STNLFFM algorithm is faster than that of the STARFM and ESTARFM algorithms, the STNLFFM algorithm is still time-consuming, and the calculation efficiency needs to be further improved.